\documentclass[letterpaper]{article}

\usepackage{natbib,alifeconf}  
\usepackage{amsmath}
\usepackage{url,hyperref,cleveref}
\usepackage{booktabs}
\usepackage{lipsum}
\usepackage{xcolor}
\usepackage{subcaption}

\newcommand\blfootnote[1]{%
  \begingroup
  \renewcommand\thefootnote{}\footnote{#1}%
  \addtocounter{footnote}{-1}%
  \endgroup
}

\newif\ifnewtext
\newtextfalse

\newcommand{\NEW}[1]{%
    \ifnewtext
        {\color{red}#1}
    \else
        #1  
    \fi
}

\title{Microcosmos: Reimagining Artificial Life for the GPU Era}

\author{
    Mark Tensen$^{1,*}$,
    Ciaran Regan$^{1,2,3}$,
    Bert Wang-Chak Chan$^{1}$, \\
    {\Large 
    Mizuki Oka$^{1,4}$,
    Kenneth O. Stanley$^{5}$, \and
    Grisha Szep$^{1,4}$}
     \\
    \mbox{}\\
    $^1$Artificial Life Institute, Japan \\
    $^2$Sakana AI, Japan \\
    $^3$University of Tsukuba, Japan \\
    $^4$Chiba Institute of Technology, Japan \\
    $^5$Lila Sciences, USA \\
    $^*$mark.tensen@alife.institute
}

\begin{document}

\maketitle

\begin{abstract}
Most artificial life simulators either operate on abstract substrates disconnected from physical reality, or simulate physically grounded worlds that do not scale to the population sizes required for open-ended evolution. We present Microcosmos, a simulation engine in which artificial lifeforms are modeled as elastic filament chains inhabiting a two-dimensional viscous fluid world, designed from the ground up for modern GPU hardware and end-to-end differentiable simulation. We validate the engine through four experiments. Hand-designed locomotion strategies confirm that the fluid coupling respects known physical constraints. Gradient-based optimization of filament folding demonstrates both the full differentiability of the simulator and the expressivity of the filament encodings. Neuroevolution and quality-diversity search produce a wide range of swimming and chemotaxis behaviors automatically. Linear scaling with particle count confirms the engine supports large-scale simulation. Microcosmos is released as an open platform with the long-term goal of supporting large-scale open-ended evolutionary simulations, designed to be physically plausible and computationally scalable.
\end{abstract}

Submission type: \textbf{Full Paper}\\

Code available at: \url{https://github.com/alife-institute/microcosmos} \\
Supplementary videos available at: \url{https://alife.institute/microcosmos-supp}

\blfootnote{\textcopyright  2026 [Mark Tensen, Ciaran Regan, Bert Wang-Chak Chan, Mizuki Oka, Kenneth O. Stanley, and Grisha Szep]. Published under a Creative Commons Attribution 4.0 International (CC BY 4.0) license.}

\begin{figure}[tp!]
\centering

\begin{subfigure}{\columnwidth}
\centering
\includegraphics[width=\columnwidth]{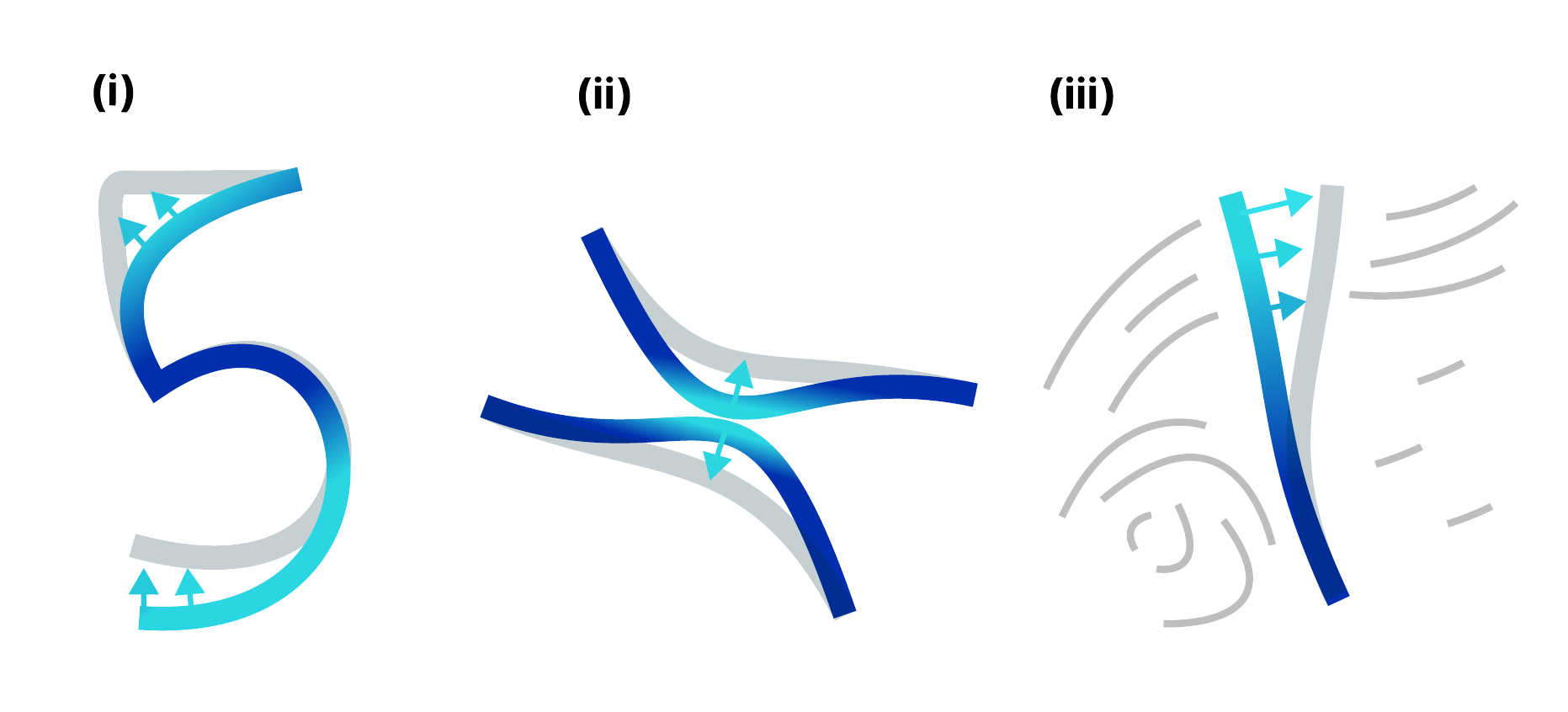}
\caption{Filaments bend, interact and flow in a fluid field.}
\label{fig:physics_components}
\end{subfigure}

\begin{subfigure}{\columnwidth}
\centering
\includegraphics[width=\columnwidth]{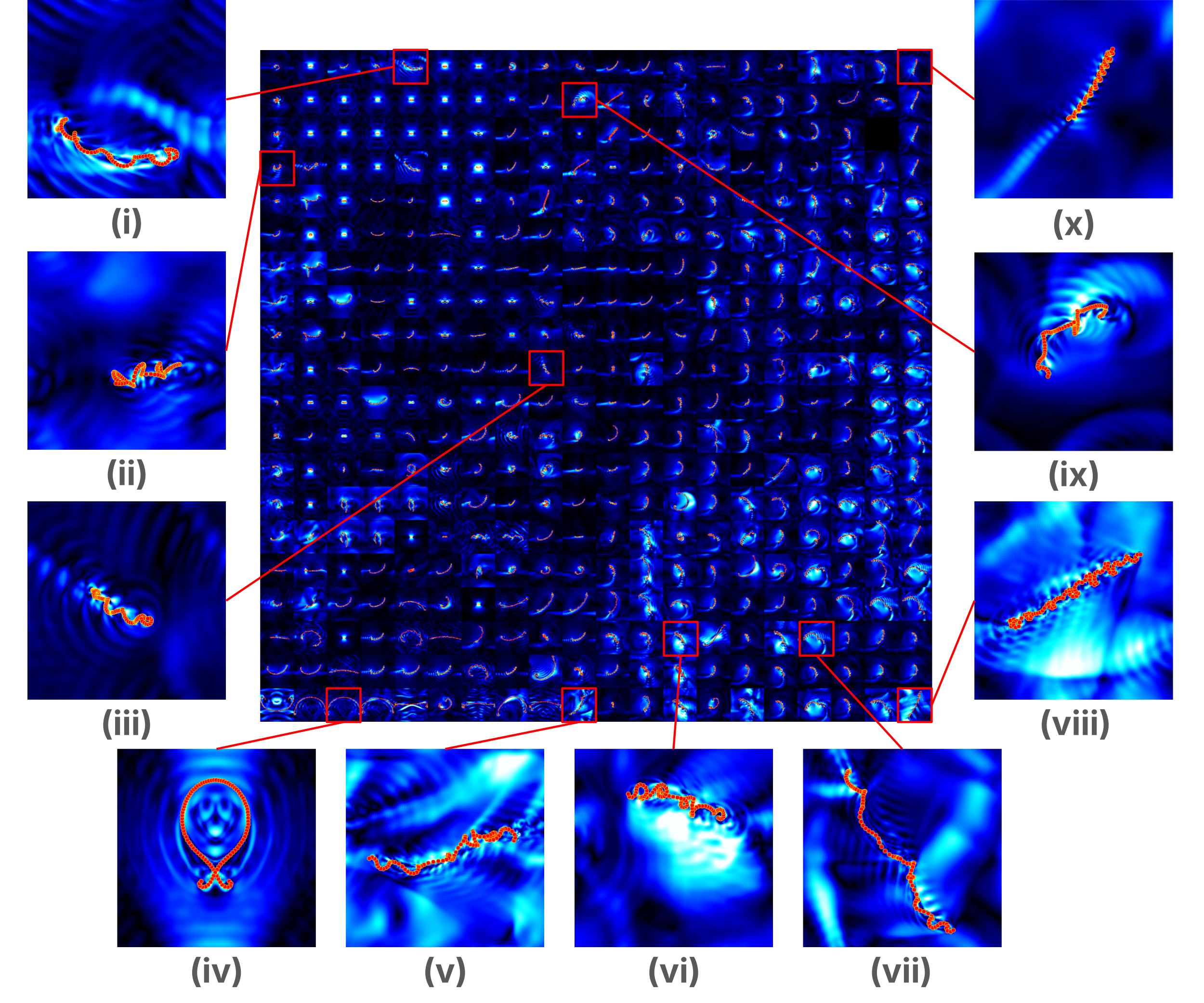}
\caption{Diverse swimming gaits discovered via QD search.}
\end{subfigure}

\caption{\textbf{Microcosmos: a scalable simulator for artificial life.} Individuals are modeled as flexible filaments in a simulated fluid environment. (a) The three core physics components: (i) Filaments have preferred resting shapes, able to bend and deform elastically. (ii) Self-avoidance and inter-body repulsion mediated by scalar fields. (iii) Two-way coupling between individuals and the surrounding fluid. (b) Diverse swimming strategies discovered by quality-diversity search, illustrating the behavioral richness the simulator supports. See supplementary materials \href{https://alife.institute/microcosmos-supp/index.html\#fig_s1a}{Figure~S1a}
for animations.}
\label{fig:overview}
\end{figure}

\section{Introduction}
What is preventing artificial life (ALife) from achieving the kind of open-ended evolutionary complexity we observe in nature? Despite decades of progress, many of the open problems identified by \citet{bedau2000open} remain unsolved. We argue that the field is held back by both a lack of physical grounding in our simulation substrates and insufficient computational scalability to support evolutionary search at scale. Abstract rule-based substrates \citep{langton1986studying,chan2019lenia,ray1991evolution,alakuijala2024computational} are efficient to evolve but disconnected from the physical principles underlying life, as studied in fields such as biophysics and fluid dynamics. Conversely, more physically grounded simulators tend to be computationally intractable for large scale evolutionary search \citep{faure2012sofa,nedelec2007collective}, operate at too high a level of abstraction~\citep{heinemann2024alien, alakuijala2024computational}, or are designed for reinforcement learning rather than the open-ended dynamics of life~\citep{bhatia2021evolution, matthews2025kinetix, lagemann2025hydrogym}. Progress demands simulators that are physically grounded enough to be credible, efficient enough to support evolutionary search at scale, and \NEW{behaviorally} rich enough to inspire.

Looking to the real world for guidance, we draw inspiration from two domains. The first is locomotion at the microscale, where organisms inhabit a world governed by viscous fluid dynamics and where inertia is negligible \citep{purcell1977life}. The diversity of swimming gaits, body plans, and collective behaviors observed in microorganisms arises from the interplay between elastic body mechanics and the surrounding fluid \citep{lauga2009hydrodynamics}. The second is the diversity of shape and function that emerges from protein folding, where linear chains of amino acids collapse into three-dimensional structures that determine biological activity. In both cases, the underlying building blocks are filaments: chains of connected units with simple nearest-neighbor topology. This shared structure is not a coincidence. Filamentous organization, from bacterial flagella to cytoskeletal networks to nematode body plans, is among the most ancient and ubiquitous forms of biological complexity \citep{elgeti2015physics}. Furthermore, nearest-neighbor connectivity makes the simulation of these filaments highly scalable.

Here we present Microcosmos, a scalable simulation engine that takes these filaments as its core abstraction for artificial life. Artificial lifeforms are modeled as elastic filament chains inhabiting a two-dimensional viscous fluid world, with genetic encodings that specify both form and behavior. Within this world, creatures swim, fold into precise shapes, and forage across chemical gradients. We validate these capabilities using a broad range of experiments. We demonstrate that the fluid simulation supports lifelike locomotion strategies, including jellyfish, cilia, and tadpole gaits, consistent with known physical constraints. We demonstrate the full differentiability of Microcosmos by training filaments to fold into target shapes via gradient descent. Through neuroevolution and quality-diversity (QD) search, we show that a wide range of swimming strategies can be discovered automatically. Finally, we highlight the scalability of our simulation, showing that runtime increases linearly with the number of particles. To our knowledge, this is the first system to combine differentiable flexible filament simulation with a resolved fluid solver for both morphology and controller optimization. We release Microcosmos as an open platform, in the hope that it serves as a foundation for the wider artificial life community to build upon. Figure~\ref{fig:overview} gives an overview of the engine physics and showcases some of the gaits discovered via QD search.

\section{Related Work}

\subsection{Artificial Life Simulators}

Artificial life research has explored a diverse range of computational substrates in pursuit of open-ended complexification. Abstract computational worlds such as cellular automata, Lenia \citep{chan2019lenia}, Tierra \citep{ray1991evolution}, Avida \citep{ofria2004avida}, BFF \citep{alakuijala2024computational}, and Chromaria \citep{soros2014identifying} show that rich evolutionary dynamics can emerge from minimal rules, while particle-based systems such as Boids \citep{reynolds1987flocks}, Particle Life \citep{ventrella2017clusters, mohr2023particlelife}, and Particle Lenia \citep{mordvintsev2022particle} demonstrate that self-organizing collective behaviors can likewise emerge from simple local interactions. However, all of these substrates remain disconnected from physical reality, limiting the embodied behaviors they can support. Physically grounded, embodied simulators \citep{sims1994evolved, miconi2008evosphere, cheney2014unshackling, kriegman2020scalable} demonstrate that collision-based worlds can produce diverse behaviors; however such approaches are often computationally intractable. Other systems \NEW{achieve tractability through GPU-accelerated agent-based frameworks \citep{richmond2010high, richmond2023flame} or architectures built for indefinite scalability \citep{ackley2014indefinitely},} but operate at higher levels of abstraction \citep{suarez2019neural, lu2024jaxlife, heinemann2024alien, whidden2025mote}. Microcosmos addresses these limitations by grounding artificial life in viscous fluid dynamics, producing embodied creatures whose behaviors are physically plausible, visually compelling and computationally tractable at scale.

\subsection{Bio-Inspiration}

The microscopic world offers a compelling substrate for artificial life research. Filamentous structures (chains of cells, polymers, flagella, and cytoskeletal networks) are among the most ancient forms of biological organization. From bacterial flagella to nematode body plans, elongated elastic bodies immersed in viscous fluids represent a minimal yet expressive design space in which evolution has produced remarkable functional diversity \citep{elgeti2015physics}.

The physics of this regime imposes fundamental constraints on what strategies can succeed. At low Reynolds numbers, the Navier-Stokes equations become time-reversible, and Purcell's scallop theorem dictates that reciprocal motion cannot produce net locomotion in Stokes flow \citep{purcell1977life}. Breaking this symmetry requires time-irreversible deformation strategies, such as the undulatory waves of \textit{C.~elegans} \citep{boyle2012gait}, propagating transverse waves analyzed by \citet{taylor1951}, or the asymmetric ciliary strokes described in \citet{purcell1977life}. These constraints make low-Reynolds-number locomotion a natural testbed for studying how physics shapes the evolution of morphology and behavior.

\subsection{Computer Graphics}

Computer graphics (CG) research is fundamentally concerned with the efficient synthesis of believable virtual environments, and breakthroughs in CG have historically driven breakthroughs in ALife. Boids \citep{reynolds1987flocks}, originally a CG technique for animating flocks, became a foundational model for emergent collective behavior. Sims' virtual creatures \citep{sims1994evolved} applied rigid-body physics from computer animation to synthetic biological evolution. This trend continues with differentiable graphics tools such as DiffTaichi \citep{hu2019difftaichi} and DiffPD \citep{du2021diffpd}, which enable gradient-based optimization of soft bodies and fluid-structure interactions, and with voxel-based physics used to design soft robots \citep{cheney2014unshackling, kriegman2020scalable}. More recently, advances in real-time physics simulations have enabled RL benchmark frameworks such as HydroGym \citep{lagemann2025hydrogym} and FluidGym \citep{becktepe2026plug}. Microcosmos continues this lineage, drawing on GPU-efficient fluid and soft-body simulations from CG to ground artificial life in the mechanics that govern life at the microscale.

\subsection{Design Optimization in Fluids}

Microcosmos sits at the intersection of three research areas: flexible filament simulation, fluid dynamics, and design optimization. Each pairwise combination has been explored. Filaments have been simulated in resolved fluids without optimization \citep{tekinalp2025self, tian2011iblbm}, but these systems are not differentiable, so any search over morphology must rely on gradient-free methods. Filament mechanics have been made differentiable \citep{bergou2008discrete, stuyck2023diffxpbd}, enabling gradient-based optimization of rest shapes and stiffness, but without hydrodynamic coupling the optimized morphologies have no notion of locomotion or drag. Differentiable fluid solvers have been paired with design optimization \citep{ma2021diffaqua, lee2023aquarium, fan2026diff}, but these systems use rigid or beam-like bodies and target engineering inverse problems rather than morphological evolution.

The closest prior systems each miss one of the three ingredients. SophT \citep{tekinalp2025self} has resolved fluid coupling with Cosserat rods but is not differentiable. PyElastica \citep{zhang2019pyelastica} provides Cosserat rods with optimization via reinforcement learning \citep{naughton2021elastica} but uses local drag approximations rather than a resolved fluid solver. Diff-FlowFSI \citep{fan2026diff} achieves full differentiability with resolved flow, but its beam elements lack the topological flexibility needed for morphological evolution. Microcosmos occupies this remaining gap: gradient-based optimization of flexible filament morphology through a fully coupled lattice Boltzmann fluid simulation, designed from the ground up for evolutionary search over body plans and behaviors.

\section{Methods}

The Microcosmos engine consists of two coupled components: \textbf{filaments}, which represent artificial lifeforms, and \textbf{fields} which represent the environment. Filaments are implemented as graphs, where each node's state describes its position in continuous space. Fields, on the other hand, are modeled as discrete grids which store information about the world, such as the fluid state~\footnote{In fluid-solid-interaction research this is a common paradigm, where grid-based fields and continuous valued graphs are called the Eulerian and Lagrangian components respectively \citep{peskin2002immersed}.}. These two elements are overlaid, interacting as a dynamical system. This formulation could be considered a middle ground between grid-based and particle-based paradigms, such as Cellular Automata and Particle Life, respectively.\footnote{It should be noted that analogs to this paradigm already exist within the field of Artificial Life, for instance in the modeling of stigmergy-based systems like slime molds. Here, agents operate in continuous space, but deposit pheromones onto a discrete environmental grid. These pheromones are subsequently diffused across the grid, where they are read by other agents to influence their behavior \citep{jones2010characteristics}.} \NEW{The entire simulator is implemented in JAX~\cite{jax2018github}, enabling GPU acceleration and end-to-end automatic differentiation.} 

Beyond biological inspirations, this filament-field formulation is largely motivated by scalability. The engine must support large populations of interacting filaments over evolutionary timescales. As such, every algorithmic choice avoids quadratic scaling with the number of nodes. For instance, the fluid solver requires only local updates. These design choices make the simulation scalable and fully parallelizable.

\subsection{Filaments}

\NEW{Filamentous structures span the micro- and nanoscale, from bacterial flagella and cytoskeletal networks that drive low-Reynolds-number locomotion, to proteins that fold one-dimensional sequences into three-dimensional volumes, to contractile muscle filaments.} We model filaments as discrete elastic rods that resist both stretching and bending \citep{bergou2008discrete}. These can be extended to have loops and branches with a graph representation. Intersections between filaments are avoided via a grid-based steric repulsion field (see Fields below) rather than per-particle collision handling. The steric field does not \NEW{guarantee} exact collision dynamics; however, the benefits of scaling are worth the loss in accuracy.

Each segment carries a per-edge orientation angle $\theta$, following the Cosserat rod formulation \citep{antman2005nonlinear}. In this framework, $\theta$ defines the material frame of the edge.
To enable local control, the physics parameters are set on the edge level instead of globally: rest length $L_{\text{rest}}$, bending rest angle $\theta_{\text{rest}}$, and optionally stiffness coefficients ($k_{\text{bend}}$, $k_{\text{shear}}$, $k_{\text{stretch}}$) controlling how strongly each constraint is enforced. Together, they define the creature's morphology and serve as the targets of both gradient-based and evolutionary optimization.

The dynamics are solved using Position-Based Dynamics (PBD) \citep{muller2007}, an iterative constraint projection method that directly modifies node positions and edge orientations to satisfy geometric constraints. PBD was chosen for two reasons: it is unconditionally stable regardless of timestep, and its operations are simple arithmetic that JAX can differentiate through directly, unlike implicit solvers that would require differentiating through a linear solve.

Enforcing Cosserat constraints within a PBD framework follows established methods \citep{umetani2014position}, which we adapt here for the planar (2D) case. Each iteration decouples the solver into two phases. First, a bending pass updates $\theta$ toward $\theta_\text{rest}$. Second, a position pass enforces nodal connectivity by resolving stretch and shear errors based on the updated orientations and rest length $L_\text{rest}$.

\paragraph{Bending Pass}
For adjacent edges with material frame angles $\theta_1$ and $\theta_2$, we define the bending constraint $C_{\text{bend}} = (\theta_2 - \theta_1) - \theta_{\text{rest}}$. Applying a bending stiffness $k_{\text{bend}}$, the material frames are updated symmetrically:

\begin{equation}
    \theta_1 \leftarrow \theta_1 + \frac{1}{2} k_{\text{bend}} \, C_{\text{bend}}, \qquad
    \theta_2 \leftarrow \theta_2 - \frac{1}{2} k_{\text{bend}} \, C_{\text{bend}}
\end{equation}

\paragraph{Position Pass}
For an edge connecting nodes $\mathbf{x}_1$ and $\mathbf{x}_2$, let $\mathbf{v} = \mathbf{x}_2 - \mathbf{x}_1$ be the edge vector and $\mathbf{d}(\theta) = [\cos \theta, \sin \theta]^T$ be the material director. The stretch constraint along the geometric tangent is:
\begin{equation}
    \mathbf{C}_{\text{stretch}} = \left( \|\mathbf{v}\| - L_{\text{rest}} \right) \frac{\mathbf{v}}{\|\mathbf{v}\|}
\end{equation}
To decouple stretch and shear, the total positional correction $\Delta \mathbf{x}$ blends this stretch constraint with the full Kirchhoff-Love error, $(\mathbf{v} - L_{\text{rest}} \mathbf{d})$, using a shear stiffness parameter $k_{\text{shear}} \in [0, 1]$:
\begin{equation}
    \Delta \mathbf{x} = \frac{1}{2} k_{\text{stretch}} \left[ \mathbf{C}_{\text{stretch}} + k_{\text{shear}} \big( \mathbf{v} - L_{\text{rest}} \mathbf{d} - \mathbf{C}_{\text{stretch}} \big) \right]
\end{equation}
The node positions are then updated symmetrically:

\begin{equation}
    \mathbf{x}_1 \leftarrow \mathbf{x}_1 + \Delta \mathbf{x}, \qquad
    \mathbf{x}_2 \leftarrow \mathbf{x}_2 - \Delta \mathbf{x}
\end{equation}

\subsection{Fields}
\label{sec:fields}

The primary field is the fluid velocity, a persistent vector field simulated using the Lattice Boltzmann Method (LBM) with the D2Q9 model \citep{qian1992lattice} (two dimensions, nine discrete velocities). LBM was chosen because each lattice site updates independently during the streaming step, making it naturally parallel on GPU ($\mathcal{O}(n)$) and compatible with spatial domain decomposition. Operating at the mesoscopic scale between molecular dynamics and direct Navier-Stokes solvers, LBM handles moving filament boundaries without remeshing (bending the grid to follow the swimmer), and its viscosity is controlled by a single relaxation parameter $\tau$, making it straightforward to sweep across physical regimes. Self-avoidance between filament nodes is handled through a grid-based steric repulsion field rather than pairwise distance checks, avoiding $\mathcal{O}(n^2)$ scaling: node positions are deposited onto a grid, diffused with a Gaussian filter via FFT convolution to produce a density field, and the gradient of this density field generates repulsive forces.

In neuroevolution experiments, we additionally introduce a persistent scalar energy field consisting of Gaussian packets placed at random positions, providing a chemotaxis target for evolving creatures.

\subsection{Fluid-filament Interaction}

Fluid and filaments interact through the Immersed Boundary Method (IBM) \cite{peskin2002immersed}. IBM enables a two-way information exchange: first, the filaments' nodes sample the local fluid velocity to determine their own drag, and second, the agent broadcasts its own velocity back onto the grid. By smoothing these interactions over a small neighborhood (a diffuse interface), the method converts single node positions 
into a continuous, approximated solid body.

The resulting diffuse (as opposed to a solid) fluid-solid boundary comes with the challenge of fluid leaking through the filaments, which is especially the case with a zero-thickness filament.
To address this issue, we used a number \NEW{of} LBM/IBM best-practices: 
Multi-Direct Forcing (MDF) \citep{wang2008combined}, Two-Relaxation-Time (TRT) \citep{ginzburg2008two}, and broadcasting the single filament into a two-layer version of itself to artificially amplify a zero-thickness filament into a solid object~\citep{mittal2005immersed}.

\subsection{Encoding Schemes and Search Methods}

Microcosmos is highly flexible in terms of the representation of filaments, supporting both direct and indirect encoding schemes \citep{fekiavc2011review, clune2011performance, miikkulainen2021biological} that map from searchable parameters (the genotype) to the physical form and behavior (the phenotype).

Direct encoding uses one-to-one mappings, where every individual parameter is specified and potentially optimized. For Microcosmos, that means the control signals and parameters ($\theta_\text{rest}, L_\text{rest}, k_\text{bend}, k_\text{shear}, k_\text{stretch}$, etc) are individually specified for each node. 

In contrast, indirect encoding uses a generative or developmental model, with its own searchable parameters, to generate the physical traits for each node based on spatial-temporal or sensory inputs. Compositional Pattern Producing Network (CPPN) \citep{stanley2007cppn}, L-systems \citep{de2011automatic} and cellular automata \citep{gutierrez2005non, najarro2022hypernca} are prominent examples. 

The Microcosmos engine is search-method agnostic and end-to-end differentiable. In our experiments, we use gradient-based optimization for filament folding experiments (Filament Folding) and QD search for neuroevolution experiments (Neuroevolution of Controller Circuit, QD Search for Locomotion Strategies). We also have locomotion experiments (Hand-designed Locomotion) that use hand-selected parameters to validate the fluid-filament coupling.

\begin{figure}[tb]
\centering
\includegraphics[trim={4cm 2cm 0 1cm}, clip, width=1.1\columnwidth]{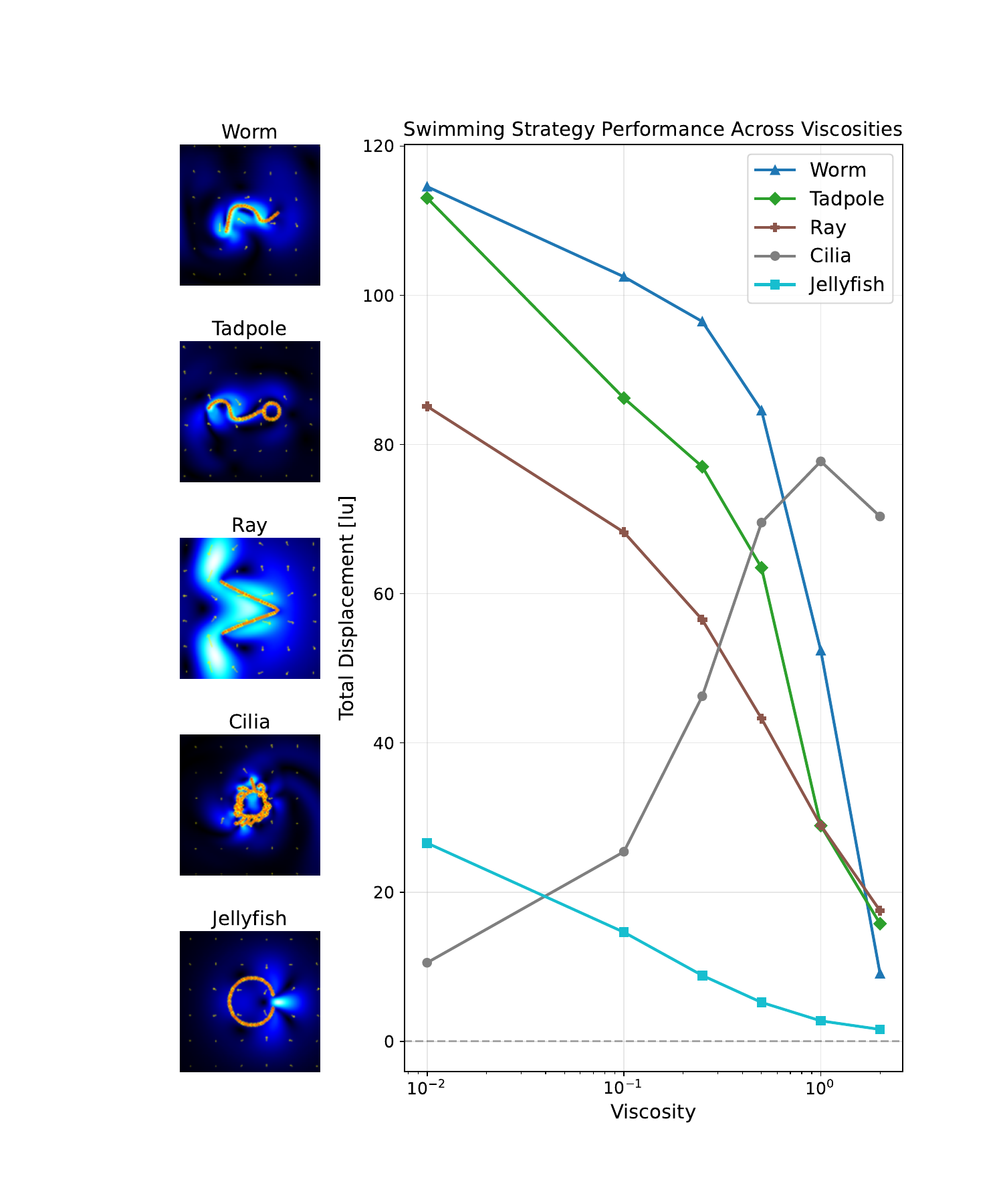}
\caption{\textbf{Realistic fluid dynamics of the Microcosmos engine} as confirmed by five creatures with hand-designed geometry and locomotion (left) and viscosity sweep (right). Consistent with Purcell's scallop theorem, time-reversible strategies (e.g. ray) produce zero net displacement at high viscosity, while non-reciprocal strategies (e.g. cilia) maintain viable locomotion. See supplementary materials \href{https://alife.institute/microcosmos-supp/index.html\#fig_s2}{Figure S2} for animations. }
\label{fig:locomotion}
\end{figure}

\begin{figure*}[tb]
\centering
\includegraphics[width=\textwidth]{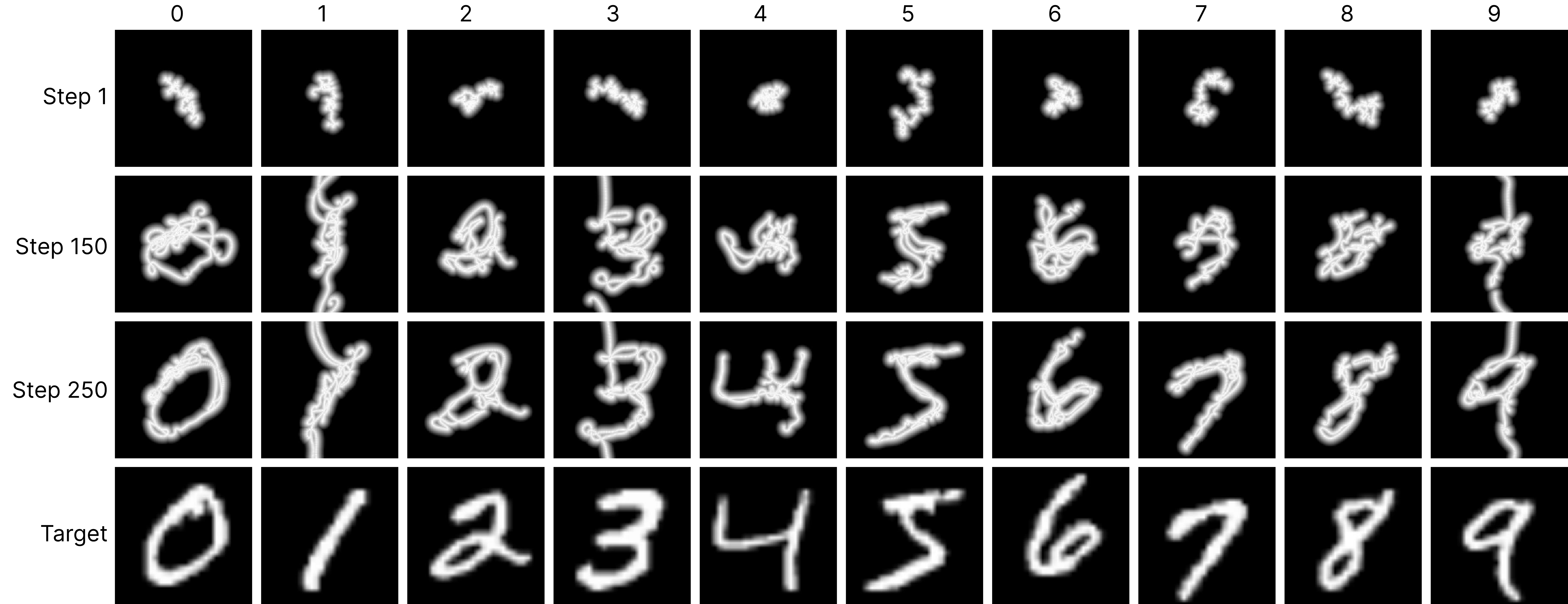}
\caption{\textbf{Differentiability of the Microcosmos engine} as demonstrated by MNIST digit folding via stochastic gradient descent (SGD). Each column shows a different target digit (0--9). The first three rows show the filament folding through a single 250-step simulation run (top to bottom: early, mid, and late), with the bottom row showing the corresponding target digit. See supplementary materials \href{https://alife.institute/microcosmos-supp/index.html\#fig_s3}{Figure S3} for animations.}
\label{fig:mnist}
\end{figure*}

\begin{figure}[tb]
\centering


\begin{subfigure}{\columnwidth}
\centering
\includegraphics[width=0.8\columnwidth]{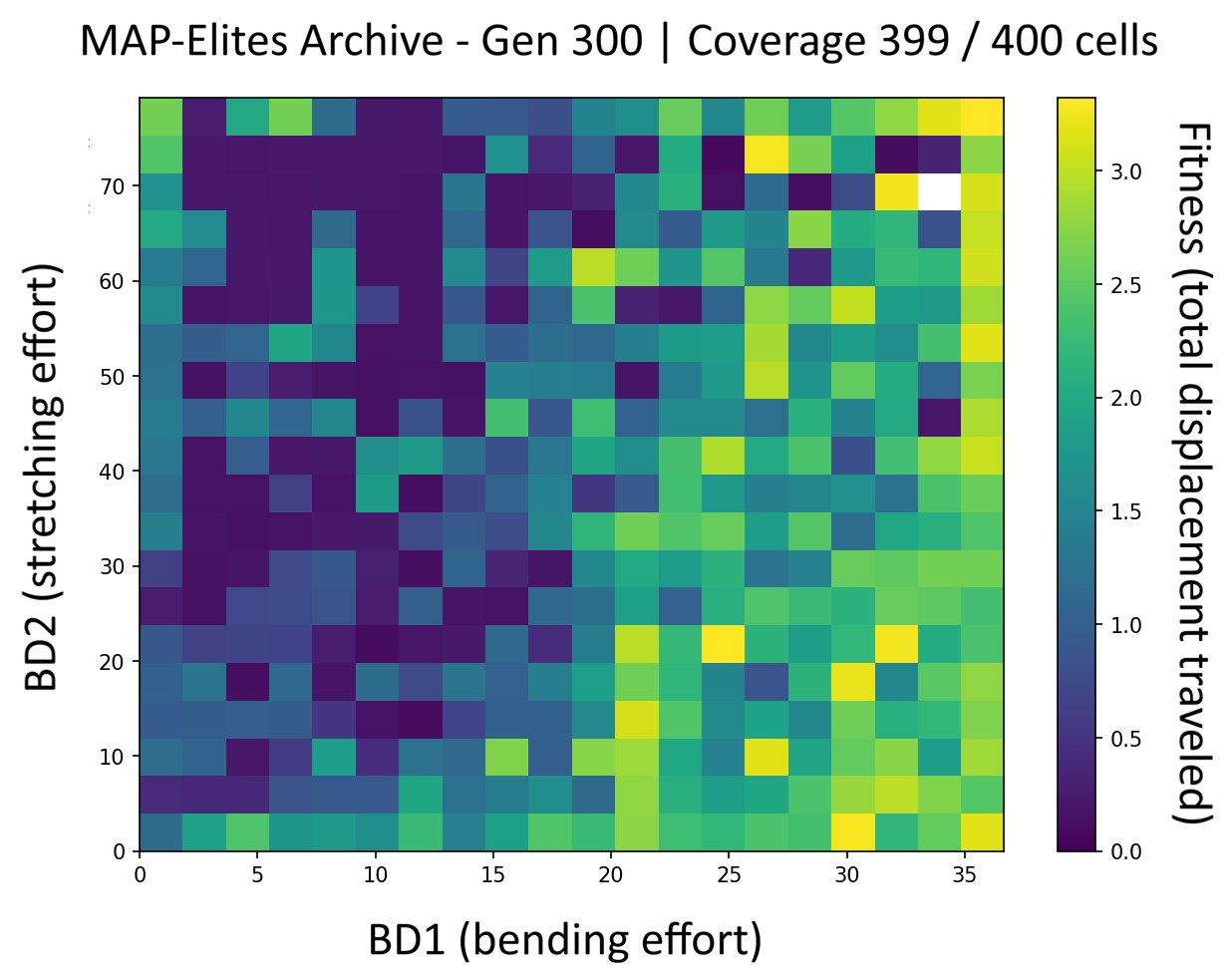}
\caption{Fitness grid of elite creatures, using the same behavioral space as in Figure~\ref{fig:overview} (b). High-fitness swimmers are widely distributed inside the grid, confirming the effectiveness of the BDs in use.}
\end{subfigure}

\begin{subfigure}{\columnwidth}
\centering
\includegraphics[width=0.9\columnwidth]{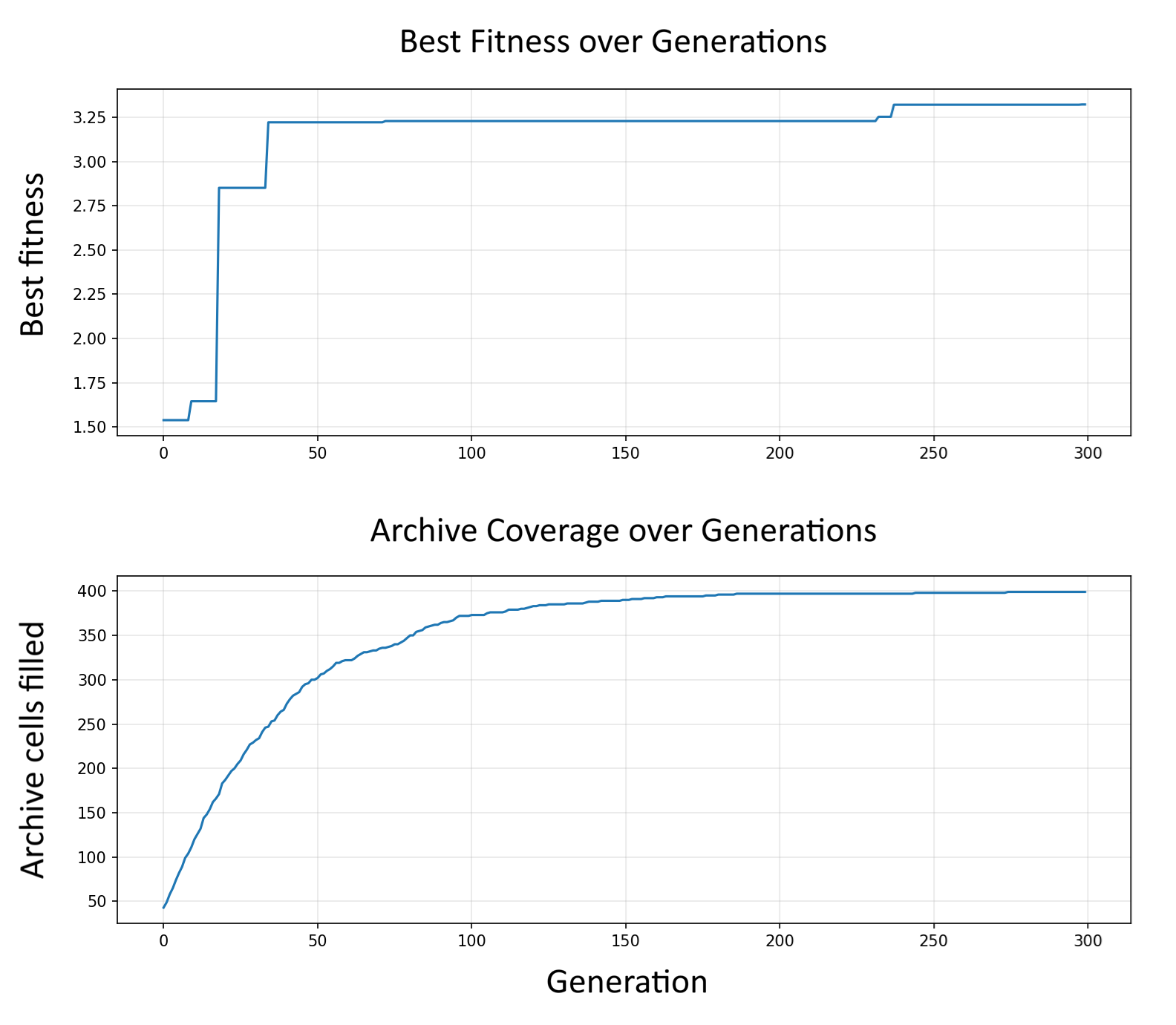}
\caption{Best fitness and archive coverage through evolution progress.}
\end{subfigure}

\caption{\textbf{Evolvability of diverse locomotion strategies} as demonstrated by QD algorithm MAP-Elites. See Figure~\ref{fig:overview} (b) for the MAP-Elites archive of elite creatures, snapshot at simulation
step 500. }
\label{fig:qd}
\end{figure}

\section{Experiments}
\label{sec:experiments}

We validate the Microcosmos engine through experiments targeting physical correctness, differentiability, evolvability, and scalability.

\subsection{Hand-designed Locomotion}
We demonstrate five hand-designed geometries and locomotion strategies that validate the physics engine's fluid coupling (Figure~\ref{fig:locomotion}). Viscosity is swept from $10^{-3}$ to $2$; this spans approximately Reynolds number $Re \sim 10^3$ to $1$ depending on the swimmer, covering the transition from inertia-dominated to viscosity-dominated swimming. Although these values already cover a useful and diverse scale, we acknowledge that outside this range our fluid-filament coupling may give rise to artifacts.

Each creature exhibits distinct behaviors: The worm and the tadpole use a similar sinusoidal wave propagation along a soft filament analogous to the undulatory waves of \textit{C.~elegans} discussed above, with the tadpole reaching lower speeds because of the drag caused by its head; the ray uses beating via a single time-dependent angle constraint; the cilia uses asymmetrical movement of short filaments attached around a stiff spherical body; the jellyfish uses contractile movement via uniform distance and bending constraints on an open circle.

The peak displacement observed in Figure~\ref{fig:locomotion} from cilia at high viscosity is a hallmark of asymmetrical swimming strokes. In these regimes, propulsive efficiency relies on the increased fluid traction generated by higher viscous resistance. Unlike the worm's undulatory waves, cilia exploit drag asymmetry more directly: a stiff stroke sweeps fluid broadside while the recovery stroke curls tightly, making them better matched to vanishingly low Reynolds numbers. Ultimately, a very high viscosity brings all movement close to a stop.

\subsection{Filament Folding}

To demonstrate the differentiability and expressivity of the simulation and filament parameterization, we optimize filaments to fold into target shapes, offering a simplified 2D analog of the protein folding problem. The angles and lengths of a 1000-node linear filament are optimized directly using Adam (learning rate $0.04$, 300 parameter update steps), targeting MNIST digit shapes. Starting from a random initial configuration, we simulate the filament for 250 simulation steps and compute the folding loss as the mean squared error between the Gaussian-splatted particle density at the final timestep and the target digit image at $128 \times 128$ resolution. Gradients of this loss flow end-to-end through the full simulation via JAX automatic differentiation, including through the PBD constraint solver and fluid computation, directly updating the morphological parameters.

Results for each MNIST digit class are shown in Figure~\ref{fig:mnist}. Gradient descent reliably recovers parameters that fold the filament into recognizable digit shapes across all ten classes. Although the reproductions are imperfect, the results confirm that the Cosserat rod formulation provides a sufficiently expressive and differentiable substrate for 2D morphogenesis.

\subsection{Neuroevolution of Controller Circuit}

We experimented with evolving a Compositional Pattern Producing Network (CPPN) \citep{stanley2007cppn} controller for the filaments. We used one shared CPPN to perform sensorimotor control for each node. The CPPN inputs include the node's position along the filament (ranging from 0.0 to 1.0) and a global timer, such that each node has a basic sense of space and time. The CPPN outputs the rest angle $\theta_\text{rest}$ and the rest length $L_\text{rest}$, such that the node can control its own movement.

Neuroevolution of Augmenting Topologies (NEAT) \citep{stanley2002neat} is used to evolve the CPPN, where the algorithm mutates the network architecture, activation functions, and connection weights. Both are implemented using TensorNEAT \citep{wang2025tensorneat}. The fitness function is defined as the total displacement traveled. We found that the highest fitness swimmers predominantly perform sinusoidal locomotion.

\NEW{We also added an ``energy'' field, a scalar field with scattered packets representing local energy akin to food. The CPPN takes the energy value at the node's position as an additional input, and fitness becomes the total displacement traveled plus the total energy collected, rewarding both movement and chemotaxis. See supplementary materials \href{https://alife.institute/microcosmos-supp/index.html\#fig_s4}{Figure~S4} for animations.}

\subsection{QD Search for Locomotion Strategies}

To discover more diverse types of locomotion, we apply QD search \citep{pugh2016qualitydiversity}. In particular, we use Multi-dimensional Archive of Phenotypic Elites (MAP-Elites) \citep{mouret2015mapelites} to search for diverse locomotion strategies. Here we show the results using \NEW{behavioral descriptors (BDs)} \NEW{defined} as the bending effort $E_{\text{bend}}$ and stretching effort $E_{\text{pos}}$, where $\theta_\text{rest}^t$ and $L_\text{rest}^t$ are rest angles and rest lengths at simulation time $t$, respectively.

\begin{align}
E_{\text{bend}} &= \sum |\theta_{\text{rest}}^{t-1} - \theta_{\text{rest}}^{t}|^2 \\
E_{\text{pos}} &= \sum |L_{\text{rest}}^{t-1} - L_{\text{rest}}^{t}|^2
\end{align}

A diverse set of swimming strategies emerged from the CPPN-QD search (Figure~\ref{fig:overview} (b)). 
The strategies can roughly be clustered into a few categories, including sinusoidal movements (i, iii, x), directional turning (v, vi, vii), complex motions (ii, viii, ix), circular structure paddling with ``flippers'' (iv). See supplementary materials \href{https://alife.institute/microcosmos-supp/index.html#fig_s1a}{Figure~S1a} and \href{https://alife.institute/microcosmos-supp/index.html#fig_s1b}{Figure~S1b} for animations.


\subsection{Scalability}
\begin{figure}
    \centering
    \includegraphics[width=0.9\linewidth]{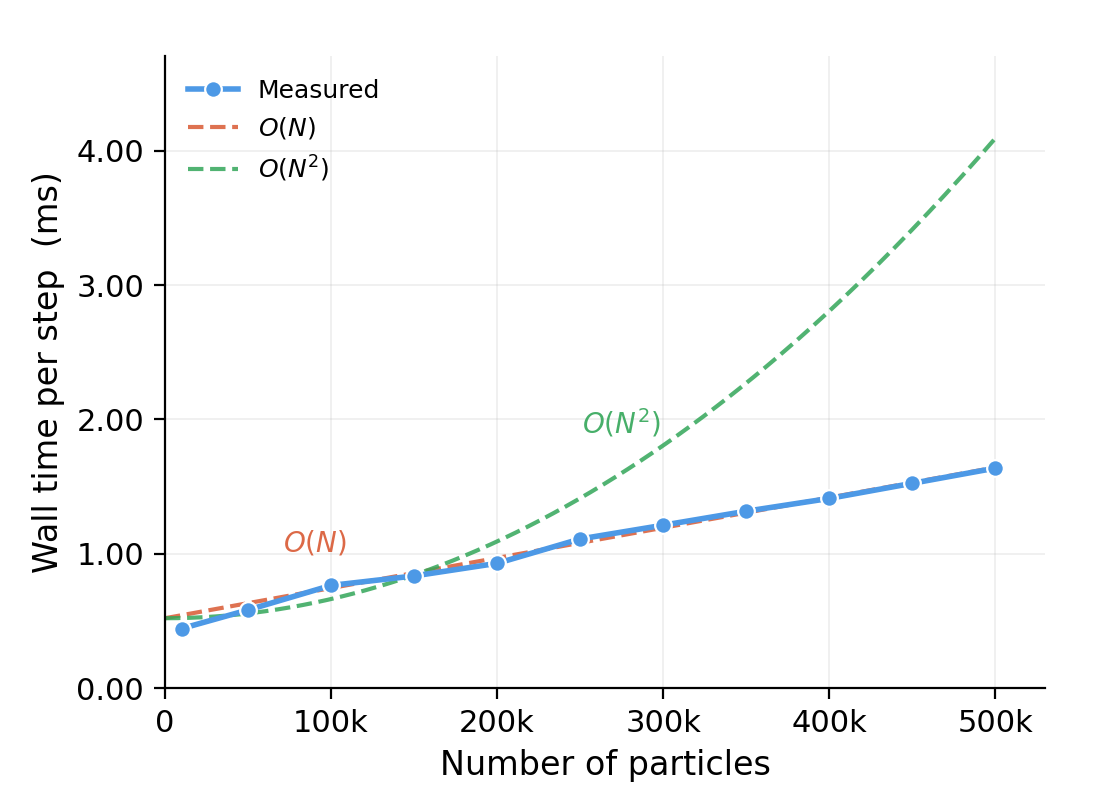}
    \caption{Wall clock time scales linearly with the number of particles in Microcosmos, in contrast to the $\mathcal{O}(n^2)$ scaling of simulators that compute pairwise particle interactions.}
    \label{fig:scaling}
\end{figure}

\NEW{To demonstrate scalability, we measure wall clock time as we increase the number of particles at a fixed grid resolution ($256\times256$), running 1000 steps with up to 500k particles\footnote{\NEW{Ran with 1 NVIDIA L40S GPU, JAX version $0.8.1$.}}. As shown in \cref{fig:scaling}, Microcosmos scales linearly, unlike simulators that rely on pairwise interactions such as Particle Life \citep{mohr2023particlelife}, which scale as $\mathcal{O}(n^2)$. Tiling strategies \citep{green2010particle} help but still degrade to quadratic scaling under non-uniform particle distributions. Every design choice in Microcosmos, from grid-based steric repulsion to the local updates of LBM, targets linear scaling.}

\section{Discussion and Conclusion}
Microcosmos was built from a simple conviction that artificial life has been held back not by a lack of imagination, but by a lack of the right tools. The simulation substrates available to the field have forced an uncomfortable choice between physical credibility and computational tractability. \NEW{We set out to narrow that gap by drawing on advances in fluid dynamics, biophysics, and computer graphics that the ALife community has largely not yet absorbed. Like any simulation, it abstracts away much of the physical world, but it occupies a useful and under-explored region between credibility and scalability.}

The result is a simulation engine in which artificial lifeforms are modeled as elastic filament chains inhabiting a two-dimensional viscous fluid world. Filaments are among the most ancient and ubiquitous structures in biology, and their nearest-neighbor topology makes them naturally scalable to simulate. The fluid environment, solved via the Lattice Boltzmann Method, imposes real physical constraints on locomotion, grounding the evolutionary dynamics in physical reality. The entire pipeline is implemented in JAX and is end-to-end differentiable, supporting both gradient-based optimization and evolutionary search.

Our experiments validate the physical correctness of the fluid coupling, the differentiability of the full simulation pipeline, and the ability to automatically discover diverse locomotion and chemotaxis behaviors, all while scaling linearly with particle count. The simulation is not without limitations. For instance, the topologies of the filaments are currently restricted to cactus graphs and the physics engine has a number of hyperparameters that may require adjustment.

We offer Microcosmos as an open platform and a call to the wider ALife community to build upon it. Our future work will focus on multi-agent interactions, not only simulating larger ecosystems but also allowing for unpredictable interactions between individuals to emerge. Our bet is that this will require some notion of birth, death and self-assembly. This would open studies of competition and symbiosis as well as the emergence of individuality and function. Further ahead, we hope to see large-scale compute runs applied to open-ended evolution in physically grounded virtual worlds. Just as scaling compute transformed modern AI, we believe scaling physically grounded simulations may do the same for artificial life.

\section{Acknowledgments}
This work was supported by JSPS KAKENHI Grant Number 24H02200 and a post-doc fellowship (PE25014). We thank Takashi Ikegami for his invaluable support.

\footnotesize
\bibliographystyle{apalike}
\bibliography{main} 

\end{document}